\documentclass{article}
\usepackage{graphicx}
\usepackage{amsfonts}
\usepackage{amsmath}
\usepackage{caption}
\usepackage{subcaption}
\usepackage{todonotes}
\usepackage{wrapfig}
\usepackage[ruled]{algorithm2e}
\usepackage{algorithmic}
\usepackage{threeparttable}
\usepackage{tabularx, booktabs}
\usepackage[final]{corl_2019}

\newcommand{\net}{S$^4$G}

\title{S$^4$G: Amodal Single-view Single-Shot $\mathbb{SE}$(3) Grasp Detection in Cluttered Scenes}

\author{
  Yuzhe Qin$^{*1}$\And Rui Chen$^{*1,2}$ \And Hao Zhu$^{1}$ \And Meng Song$^{1}$ \\
  \And Jing Xu$^{2}$ \And Hao Su$^{1}$ \\
  $^1$ University of California, San Diego \\
  \texttt{\{y1qin, haozhu, mengsong, haosu\}@eng.ucsd.edu}\\
  $^2$ Tsinghua University \\
  \texttt{chenr17@mails.tsinghua.edu.cn, jingxu@tsinghua.edu.cn}\\
}

\begin{document}
\maketitle


\begin{abstract}
Grasping is among the most fundamental and long-lasting problems in robotics study. This paper studies the problem of 6-DoF(degree of freedom) grasping by a parallel gripper in a cluttered scene captured using a commodity depth sensor from a single viewpoint. We address the problem in a learning-based framework. At the high level, we rely on a single-shot grasp proposal network, trained with synthetic data and tested in real-world scenarios. Our single-shot neural network architecture can predict amodal grasp proposal efficiently and effectively. Our training data synthesis pipeline can generate scenes of complex object configuration and leverage an innovative gripper contact model to create dense and high-quality grasp annotations. Experiments in synthetic and real environments have demonstrated that the proposed approach can outperform state-of-the-arts by a large margin. 


\end{abstract}



\section{Introduction}

Grasping is among the most fundamental and long-lasting problems in robotics study. While classical model-based methods using mechanical analysis tools~\citep{bohg2013data, graspit, dang2012semantic} can already grasp objects of known geometry, it remains an open problem of how to grasp generic objects in complex scenes. 

Recently, data-driven approaches have shed light to addressing the generic grasp problem using machine learning tools \citep{levine2018learning, jang2018grasp2vec, watkins2019multi, quillen2018deep}. In order to readily generalize to unseen objects and layouts, a large body of recent works have focused on solving 3/4 DoF(degree of freedom) grasping, where the gripper is forced to approach objects from above vertically \citep{lenz2015deep, kumra2017robotic}. Although this has greatly simplified the problem for picking and placing tasks, it has also inevitably restricted ways to interact with objects. 
For example, such grasping is unable to grab a horizontally placed plate. 
Worse still, top-down grasping often encounters difficulties in cluttered scenes with casually heaped objects, which requires extra hand freedoms for grasping buried objects. The limitation of 3/4 DoF grippers thus motivates the study of 6-DoF grippers to approach the object from arbitrary directions. We note that 6-DoF end-effector 
is essential to allow dexterous object manipulation tasks \citep{ten2017grasp, gualtieri2018learning}. 

This paper studies the 6-DoF grasping problem in a realistic yet challenging setting, assuming that a set of household objects from unknown categories are casually scattered on a table. A commodity depth camera is mounted with a fixed pose to capture this scene from only a single viewpoint, which gives a partial point cloud of the scene. The grasp is performed by a parallel gripper.

The setting is highly challenging for both perception and planning: First, the scene clutters limit viable grasp poses and may even fail the motion planning algorithms to achieve certain grasps. This challenge keeps us from considering 3/4-DoF grasp detection and restricts us to the more powerful yet sophisticated 6-DoF detection approach. Second, we make no assumptions of object categories. This \emph{open set} setting puts us in a different category from existing semantic grasping method, such as DexNet~\citep{dexnet4}. We require higher-level of generalizability based on better representation of the perceived content. Most existing methods can only work in simpler scenarios, by introducing high-quality and expensive 3D sensors for accurate scene capturing, or sensing the complete environment with multiple cameras\citep{ten2017grasp}, or assuming a scene of only a single object\citep{johns2016deep}. This challenge demands that our grasp detection has to be noise-resistant and \emph{amodal}, i.e., being able to make an educated guess of the viable grasp from only a partial point cloud. 
\begin{figure}[t!]
\setlength{\belowcaptionskip}{-0.5cm}
\centering
\includegraphics[width=\linewidth]{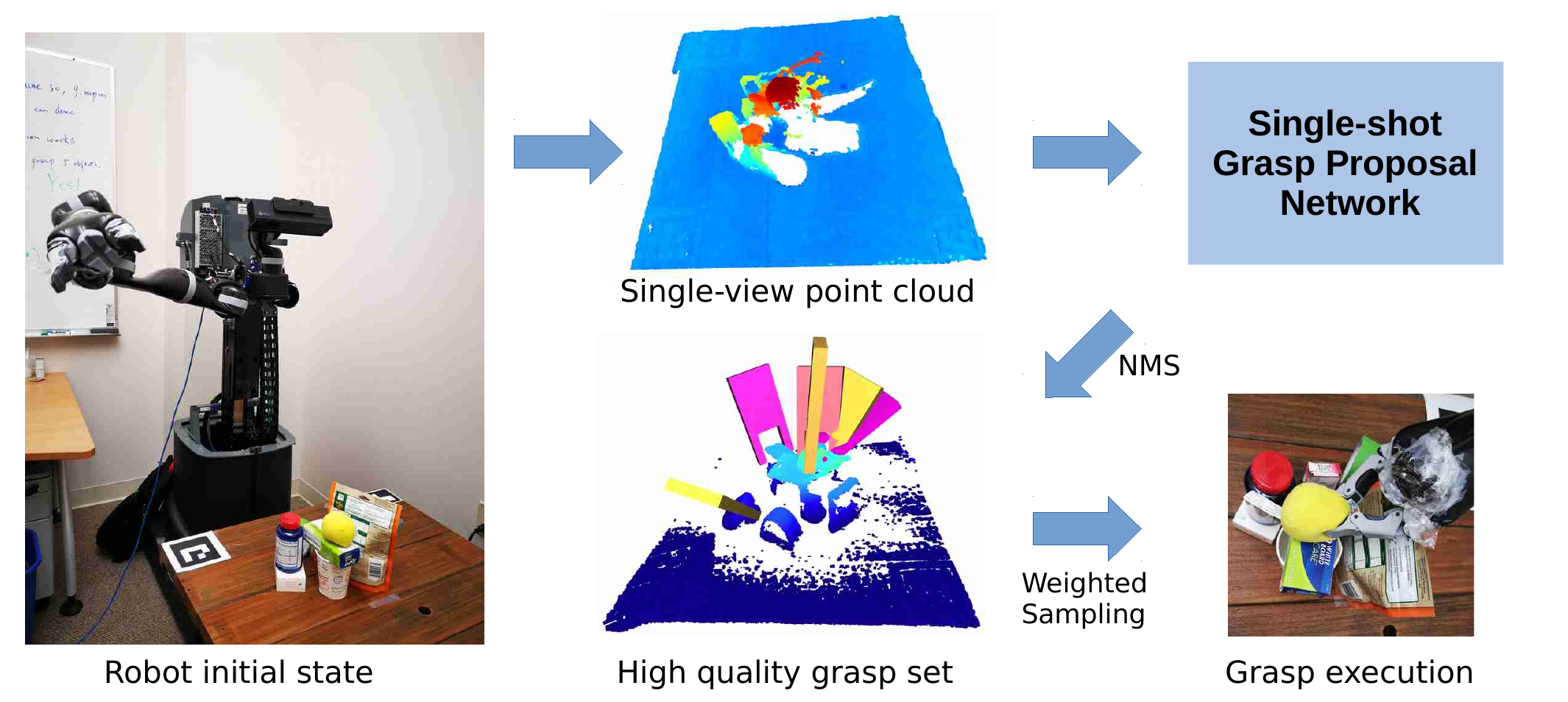}
\caption{Illustration of the pipeline of Single-Shot SE(3) Grasp Detection (\net). Taking as input the view point cloud from the depth sensor, \net~regresses the 6-DoF grasp pose directly and predicts the grasp quality for each point, which is more robust and effective.}
\label{fig:pipeline}
\end{figure}

We address the challenges in a learning-based framework. At the high level, we rely on a single-shot grasp proposal network, trained with synthetic data and tested in real-world scenarios. Our design involves (1) a single-shot neural network architecture for amodal grasp proposal; and (2) a scene-level training data synthesis pipeline leveraging an innovative gripper contact model.

By its single-shot nature, our grasp proposal network enjoys better efficiency and accuracy compared with existing deep networks in the 6-DoF grasping literature. Existing work, such as \cite{ten2017grasp}, samples grasp candidates from $\mathbb{SE}(3)$ following some heuristics and assess their quality using networks. However, the running time goes up quickly as the number of sampled grasps increases, which makes the grasp optimization too slow. Unlike these approaches, we propose to directly \emph{regress} 6-DoF grasps from the entire scene point cloud in one pass. Specifically, We are the first to propose a per-point scoring and pose regression method for 6-DoF grasp.

3D data from low-cost commercial depth sensors are partial, noisy and corrupted. To handle the imperfection of input 3D data, \net~is trained by hallucinated point clouds of similar patterns, and it learns to extract robust features for grasp prediction from the corrupted data. We propose a simple yet effective gripper contact model to generate good grasps and associate these grasps to the point cloud. At inference time, we select high quality grasps based on the proposals of the network. Note that we are the first to generate a synthetic scene of many objects, rather than a single object, in the 6-DoF grasping literature.

The core novel insight of our \net~is that we learn to propose possible grasps in this space by regression. We believe learning to regress grasp proposals would be the trend: For another problem of similar setting, object detection, the community has evolved from sliding windows to learning to generate object proposals. A second novelty is that, instead of generating training data by scenes of only a single object, we include multiple objects in the scene, with grasp proposals analyzed using a gripper contact model that considers touching area shape and size. Supplementary videos and code are at \url{https://sites.google.com/view/s4ggrapsing}.

\section{Related work}

\paragraph{Deep Learning based Grasping Methods}
\citet{caldera2018review} gave a thorough survey of deep learning methods for robotic grasping, which demonstrates the effectiveness of deep learning on this task. In our paper, we focus on the problem of 6-DoF grasp proposal. \citet{collet2011moped, zeng2017multi, mousavian20196} tackled this problem by fitting the object model to the scan point cloud to retrieve the 6-DoF pose. Although it has shown promising results in industrial applications, the feasibility is limited in generic robotic application scenarios, e.g. house-holding robots, where the exact 3D models of numerous objects are not accessible. \citet{ten2017grasp} proposed to generate grasp hypotheses only based on local geometry prior and attained better generalizability on novel objects, which was further extended by~\citet{liang2019pointnetgpd} by replacing multi-view projection features with direct point cloud representation. Because potential viable 6-DoF grasp poses are infinite, these methods guide the sampling process by constructing a Darboux frame aligned with the estimated surface normal and principal curvature and searching in its 6D neighbourhood. However, they may fail finding feasible grasps for thin structures, such as plates or bowls, where computing normals analytically from partial and noisy observation is challenging. In contrast to these sampling approaches, our framework is a single-shot grasp proposal framework~\citep{chu2018real, caldera2018review}--a direct regression approach for predicting viable grasp frames--which could handle flawed input well due to the network's knowledge. Moreover, by jointly analyzing local and global geometry information, our method not only considers the object of interest, but also its surroundings, which allows the generation of collision-free grasps in dense clutters. 

\paragraph{Training Data Synthesis for Grasping}
Deep learning methods require an enormous volume of labelled data for the training process~\citep{kumra2017robotic}, however manually annotating 6-DoF grasp poses is not practical. Therefore, analytic grasp synthesis~\citep{bicchi2000robotic} is indispensable for ground truth data generation. These advanced models have provided guaranteed measurements of grasp properties with the availability of complete and precise geometric models of objects. In practice, the observation from sensors are partial and noisy, which undermines the metric accuracy. In the service of our single-shot grasp detection framework, we first use analytic methods to generate viable grasps for each single object, and reject unfeasible grasps in densely clutter scenes. To the best of our knowledge, the dataset we generated is the first large-scale synthetic 6-DoF grasp dataset for dense clutters.

\paragraph{Deep Learning on 3D Data}

\citet{qi2016pointnet, qi2017pointnet++} proposed PointNet and PointNet++, a novel 3D deep learning network architecture capable of extracting useful representations from 3D point clouds. Compared with other architectures~\citep{maturana2015voxnet, qi2016volumetric}, PointNets are robust to varying sampling densities, which is important to real robotic applications. In this paper, we utilize PointNet++ as the backbone of our single-shot grasp detection and demonstrate its effectiveness.

\section{Problem Setting}
We denote the single-view point cloud by $\mathbf{P}$ and the gripper description by $\mathbf{G}$. A parallel gripper can be parameterized by the frame whose origin lies at the middle of the line segment connecting two figure tips and orientation aligns with the gripper axes. We therefore denote a grasp configuration as $c=(\mathbf{h}, s_{\mathbf{h}})$, where $\mathbf{h} \in \mathbb{SE}(3)$ and $s_{\mathbf{h}} \in \mathbb{R}$ is a score measuring the quality of $\mathbf{h}$.

\begin{figure}[t!]
\centering
\setlength{\belowcaptionskip}{-0.6cm}
\vspace{-0.4cm}
\includegraphics[width=0.6\textwidth]{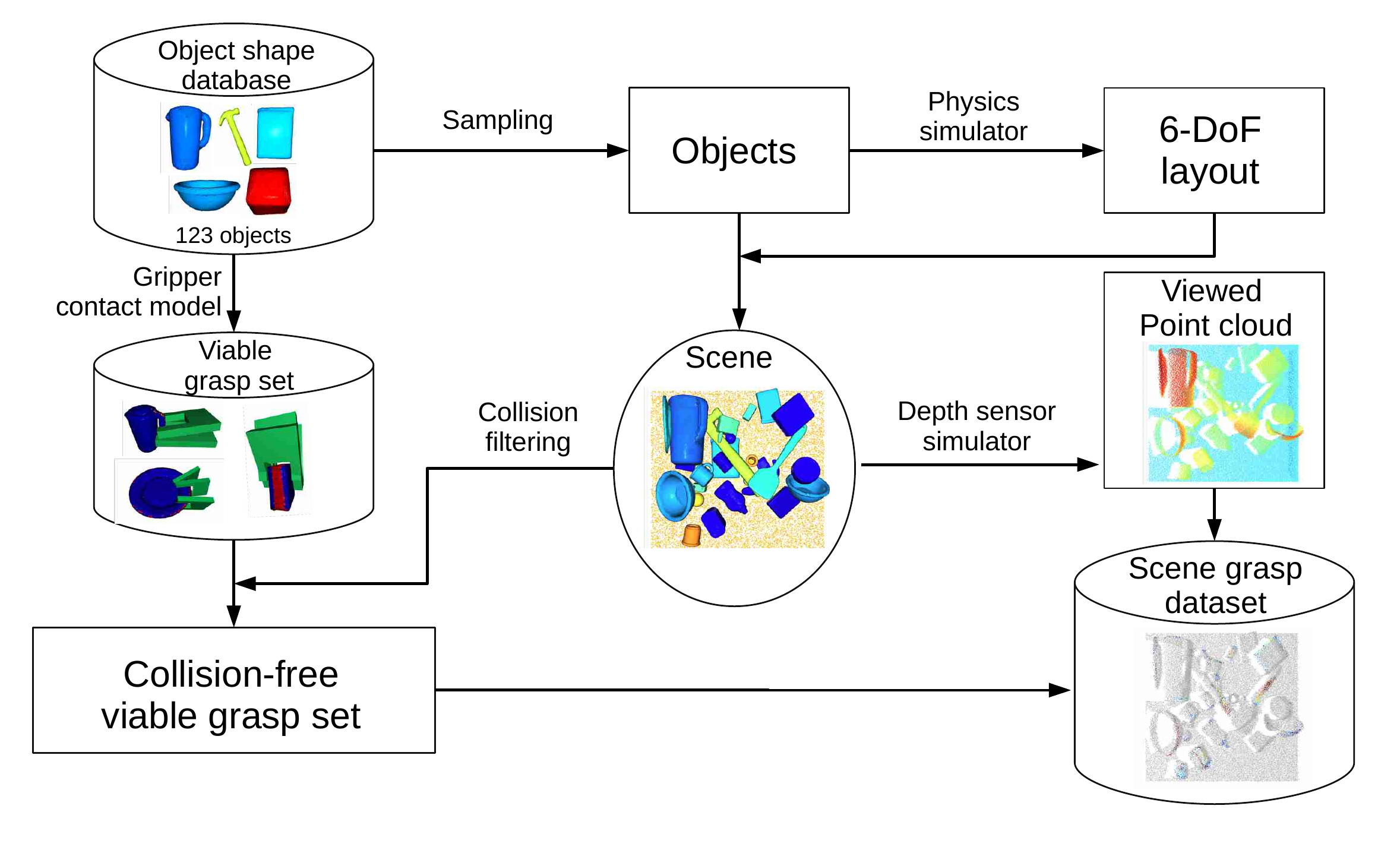}
\caption{Flowchart of scene grasp dataset generation.}
\label{fig:data_generation}
\end{figure}

\section{Training Data Generation}
\label{sec:data_generation}

To train our \net, a large scale dataset capturing cluttered scenes, with viable grasps and quality scores as groundtruth, is indispensable. Fig.~\ref{fig:data_generation} illustrates the training data generation pipeline. We use the YCB object dataset~\citep{calli2015ycb} for our data generation. Since \net~directly takes a single-view point cloud from the depth sensor as input and outputs collision-free grasps in a densely-cluttered environment, we need to generate such scenario with complete scene point cloud and corresponding partially observed point cloud. Each point in the point cloud is assigned with serval grasps which will be introduced in Sec~\ref{sec:scene_analysis} and each ground truth grasp has a $\mathbb{SE}(3)$ pose, an antipodal score, a collision score, an occupancy score, and a robustness score, which we will introduce later. On the other hand, the scene point cloud does not interact with the network explicitly, but it serves as a reference to evaluate grasps in the point cloud.


\subsection{Gripper Contact Model}



\begin{wrapfigure}{r}{0.35\textwidth}
\includegraphics[width=0.35\textwidth]{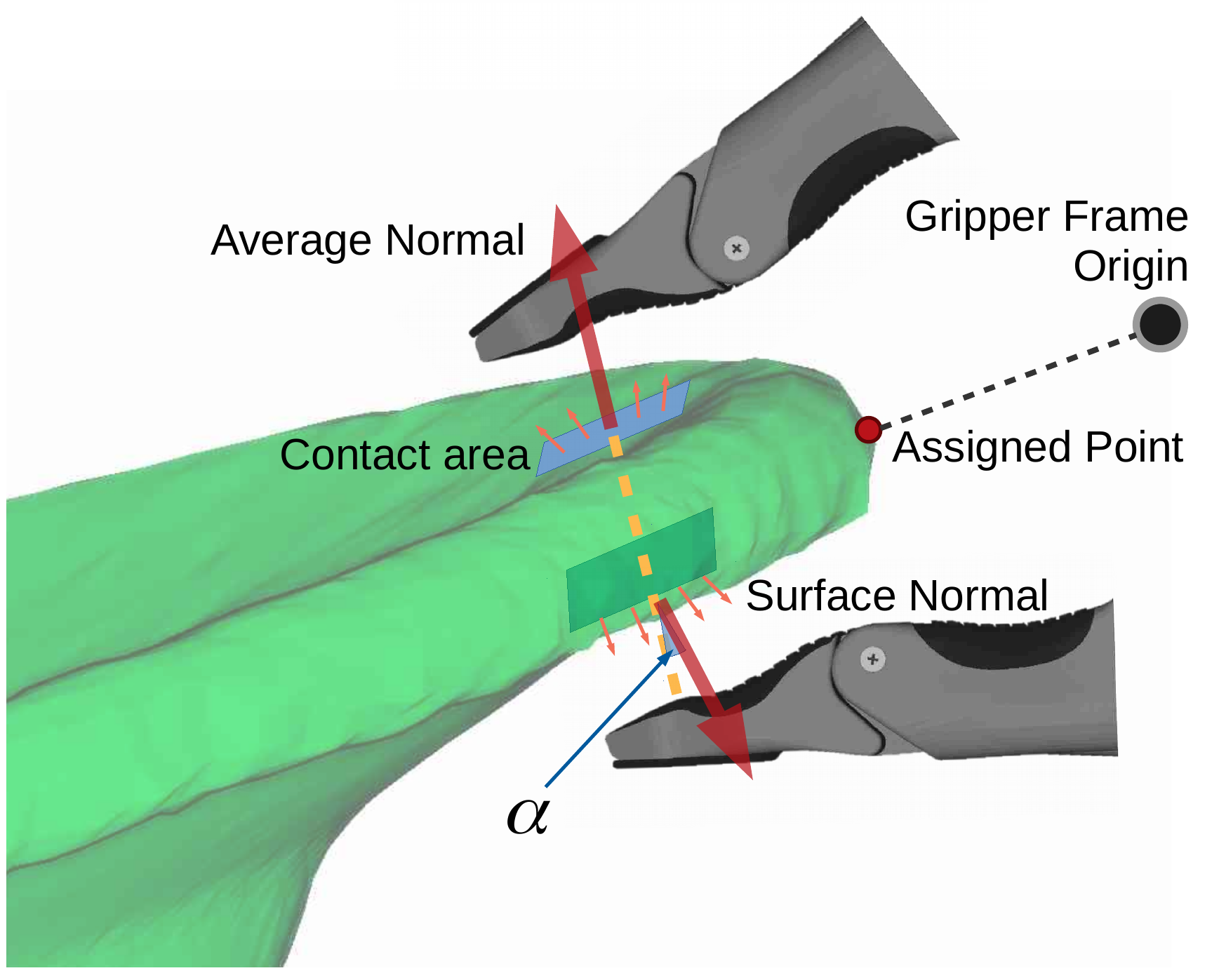}
\caption{Illustration of Gripper Contact Model}
\label{fig:contact_model}
\end{wrapfigure}

Vast literature exists to find regions suitable to grasp by analyzing the 3D geometry~\cite{sahbani2012overview}. Among these methods, force closure has been widely used to synthesize grasps and can be reduced into calculating angles between face normals, known as antipodal grasp \cite{nguyen1988constructing, chen1993finding}. Here we introduce our gripper contact model based on force closure analysis to find feasible grasps.


To be more specific, we first detect all possible contact pairs with high antipodal score $s_{\mathbf{h}}^a = \mathbf{cos}(\alpha_1)\mathbf{cos}(\alpha_2)$, where $\alpha_i$ is the angle between the outward normal and the line connecting two contact points. As illustrated in Fig.~\ref{fig:contact_model}, for each contact pair ($p_{i}$, $p_{j}$), the normal $\mathbf{n}_{i}$ at point $p_{i}$ is smoothed with radius $r$mm. Note that this step is important to grasp objects of rugged surface with high-frequency normal variation. However, we do not directly use a ball query to query its neighbors, which will lead to undesirable results at corners and edges. Instead, we remove the neighbors which has a distance along the normal direction ( calculated as $r_i^k=|(\mathbf{p}_i^k-\mathbf{p}_i)\cdot \frac{\mathbf{n}_i}{|\mathbf{n}_i|}|$) larger than $3$mm in the query ball of radius $r$ for normal calculation, where $\mathbf{p}_i^k$ is the $k$-th neighbor of point $i$. 

These two hyper-parameters have definite physical meaning, which is distinct from the approach to obtain the gripper contact model hyper-parameters in GPD~\citep{ten2017grasp} through extensive parameter tuning. As shown in Fig.~\ref{fig:contact_model}, our gripper will only interact with the object by its soft rubber pad, which allows deformation within $3$mm. And the normal smoothing radius is set as the gripper width $r=23$mm. 

In fact, our gripper model has clear advantage over Darboux frame based methods, especially at rugged surfaces and flat surfaces. For rugged surfaces, there is no principled way to decide the radius for normal smoothing, since the radius is not only relevant to the gripper, but also to the object to grasp. For flat surfaces, the principal curvature directions are under-determined. In practice, we do observe issues for these cases. For example, for plates and mugs, Darboux frame based method will likely to fail in generating a successful grasp pose for the thin wall. 

Besides the direction of contact force, we also consider the stability of the grasp. The occupancy score $s_{\mathbf{h}}^o$, which represents the volume of object within the gripper closing region $\mathbf{R(c)}$, is calculated by
\begin{equation} \label{anti}
   s^{o}_{\mathbf{h}} = \min \{ ln(| \mathbf{P_{close}}|), 6 \} ,\quad \mathbf{P_{close}} = \mathbf{R(c)} \cap \mathbf{P},
\end{equation}
where $\mathbf{P_{close}}$ is the number of points within closing region. If $s^{o}_{\mathbf{h}}$ is small, the gripper contact analysis will be unreliable. To make sure that the point cloud occupancy can correctly represent the volume, we down-sample the point cloud using voxel grid filter with a leaf size of $5$mm. 

\subsection{Physically-plausible Scene Synthesis from Objects}

Since our network is trained on synthesis data and directly applied to real world scenarios, it is necessary to generate training data closer to reality both physically and visually.

We need physically-plausible layouts of various scenes where each object should be in equilibrium under gravity and contact force. Therefore, we adopt MuJoCo engine~\cite{todorov2012mujoco} and V-HACD\citep{vhacd} to generate scenes where each object is in equilibrium. Objects initialized with random elevation and poses fall onto a table in the simulator and converge to static equilibrium due to friction. We record the poses and positions of objects and reconstruct the 3D scene.(Fig.~\ref{fig:data_generation})

Beside scene point cloud, we also need to generate viewed point clouds that will feed into the neural network. To simulate the noise of depth sensor, we apply a noise model on the distance from camera optical center to each point as $\Tilde{D}_{o,p} = (1+\mathcal{N}(0,\sigma^2))D_{o,p}$,
where $D_{o,p}$ is the noiseless distance captured by a ray tracer and $\Tilde{D}_{o,p}$ is the distance used to generate viewed point clouds. We employ $\sigma=0.003$ in this paper.

\subsection{Robustness Grasp Generation by Scene Analysis}
\label{sec:scene_analysis}

Given the scene point cloud, we can do collision detection for each grasp configurations. Collision score $s_{\mathbf{h}}^c$ is a scene-specific boolean mask indicating the occurrence of collision between the proposed gripper pose and the complete scene. As shown in our experiment, our network can better predict collision with invisible parts.

It is a common case that robot end-effector can not move precisely to a given pose due to sensor noise, hand-eye calibration error and mechanical-transmission noise. To perform a successful grasp under imperfect condition, the proposal grasp should be robust enough against gripper's pose uncertainty. In this paper, we add a small perturbation to the $\mathbb{SE}(3)$ grasp pose and evaluate the antipodal score, occupancy score and collision score for the perturbed pose. The final scalar score of each grasp can be derived as:
\begin{equation}
    s_{\mathbf{h}} = \min_j [s^a_{\mathbf{h}_j} s^o_{\mathbf{h}_j} s^c_{\mathbf{h}_j}], \quad {\mathbf{h}_j}=\mathbf{exp(\hat{\xi})}\textbf{h},
\end{equation}
where $\mathbf{\hat{\xi}} \in \mathfrak{se}(3)$ is the pose perturbation and $ \mathbf{exp}$ is the exponential mapping. The final viewed point cloud with ground truth grasps and scores will serve as training data for our \net.

\section{Single-Shot Grasp Generation}
\label{sec:network}
\subsection{PointNet++ based Grasp Proposal}
\begin{figure}[tbp]
\setlength{\belowcaptionskip}{-0.5cm}
\centering
    \includegraphics[width=\linewidth]{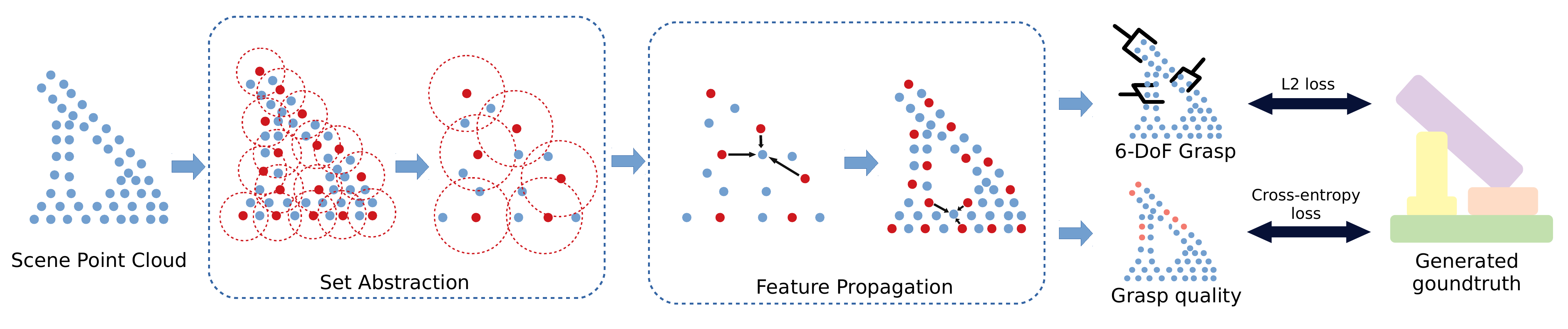}
    \caption{Architecture of Single-Shot Grasp Proposal Network based on PointNet++~\citep{qi2017pointnet++}. Given the scene point cloud, our network first extracts hierarchical point set features by progressively encoding points in larger local regions; then the network propagates the point set features to all the original points using inverse distance interpolation and skip links; finally it predicts one 6-DoF grasp pose $\textbf{h}_i$, and one grasp quality score $s_{\textbf{h}_i}$ of every point. }
    \label{fig:architecture}
\end{figure}
We design the single-shot grasp proposal network based on the segmentation version of PointNet++, which has demonstrated state-of-the-art accuracy and strong robustness over clutter, corruption, non-uniform point density \cite{qi2017pointnet++}, and adversarial attacks \cite{liu2019extending}. 

Figure.~\ref{fig:architecture} demonstrates the architecture of \net, which takes the single-view point cloud as input, and assigns each point two attributes. The first attribute is a good grasp (if exists) associated to the point by inverse indexing, and the second attribute is the quality score of the stored grasp. The generation of the grasp and quality score can be found in Sec.~\ref{sec:scene_analysis}.

The hierarchical architecture not only allows us to extract local features and predict reasonable local frames when the observation is partial and noisy, but also combines local and global features to effectively infer the geometry relationship between objects in the scene.

Compared with sampling and grasp classification~\cite{ten2017grasp, liang2019pointnetgpd}, the single-shot 6-DoF grasp direct regression task is more challenging for networks to learn, because widely adopted rotation representations such as quaternions and Euler angles are discontinuous. In this paper, we use a 6D representation of the 3D rotation matrix because of its continuity~\citep{zhou2018continuity}: for every $\mathbf{R} \in \mathbb{SO}(3)$, it is represented by $\mathbf{a} = \left[ {\mathbf{a}_1}, {\mathbf{a}_2} \right]$, $\mathbf{a}_1,\mathbf{a}_2 \in \mathbb{R}^3$, such that the mapping $f:\mathbf{a}\rightarrow\mathbf{R}$ is
\begin{equation}
\begin{array}{l}
\mathbf{R}=\left[\mathbf{b}_1, \mathbf{b}_2, \mathbf{b}_3\right]\\
\mathbf{b}_1=N\left(\mathbf{a}_1\right)\\
\mathbf{b}_2=N\left(\mathbf{a}-\left\langle \mathbf{a}_2,\mathbf{b}_1\right\rangle \mathbf{b}_1\right)\\
\mathbf{b}_3=\mathbf{b}_1 \times \mathbf{b}_2,
\end{array}
\end{equation}
where $N()$ denotes the normalization function. Because the gripper is symmetric with respect to rotation around the $x$ axis, we use a loss function which handles the ambiguity by considering both correct rotation matrices as ground truth options. Given the groundtruth rotation matrix $\mathbf{R}_{GT}$, we define the rotation loss function $L_{rot}$ as
\begin{equation}
\begin{array}{l}
L_{rot} = \underset{i \in \left\{0, 1\right\}}{\textrm{min}}\|f(\mathbf{a}_{pred}) - \mathbf{R}^{(i)}_{GT}\|^2  \\
\mathbf{R}^{(i)}_{GT} = \mathbf{R}_{GT} \begin{bmatrix}1&0&0\\0&\textrm{cos}(\pi i)&0\\0&0&\textrm{cos}(\pi i)\end{bmatrix}
\end{array}
\end{equation}

The prediction of translation vectors is treated as a regression task and the $L_2$ loss is applied. By dividing the groundtruth score into multiple levels, the grasp quality score prediction is treated as a multi-class classification task, and a weighted cross-entropy loss is applied to handle the unbalance between positive and negative data. 
We only supervise the pose prediction for those points assigned with viable grasps and the total loss is defined as:
\begin{equation}
L = \sum_{\mathbf{P}_v}{\left(\lambda_{rot} \cdot L_{rot}  + \lambda_{t} \cdot L_{t} \right)} + \sum_{\mathbf{P}_{s}}{\left(\lambda_{s} \cdot L_{s}\right)},
\label{eq:loss}    
\end{equation}
where $\mathbf{P}_v, \mathbf{P}_s$ represent the point set with viable grasps and the whole scene point cloud, respectively. $\lambda_{rot}, \lambda_{t}, \lambda_{s}$ are set to 5.0, 20.0, 1.0 in experiments.

\subsection{Non-maximum Suppression and Grasp Sampling}
\label{sec:grasp_generation}

Algorithm.~\ref{alg:grasp_selection} describes the strategy to choose one grasp execution $\textbf{h}$ from the network prediction $\mathcal C$.
\begin{wrapfigure}[20]{R}{0.5\textwidth} 
\begin{footnotesize}
  \begin{algorithm}[H]                
    \SetCustomAlgoRuledWidth{0.45\textwidth}  
    \caption{NMS and Grasp sampling}
    \textbf{Input: } Prediction $\mathcal C$: $\left\{ (\textbf{h}_i, s_{\textbf{h}_i})\right\}$ \\
    \textbf{Export: } Grasp Execution: $\textit{h}$ \\
  \begin{algorithmic}
  \STATE Executable Grasps $\mathcal H = \{\}$
  \STATE {$Sort$ $\left\{ (\textbf{h}_i, s_{\textbf{h}_i}\right\}$ $by$ $s_{\textbf{h}_i}$}
  \STATE {$i = 0$}
  \WHILE{$\mathcal Length(\mathcal H) < N$}
  \IF {$(Collision == False)$ and $\underset{min}{\textbf{h}_k \in \mathcal H}dist(\textbf{h}_i, \textbf{h}_k) > \epsilon$}
  \STATE {$Add$ ${ (\textbf{h}_i, s_{\textbf{h}_i})}$ to ${\mathcal H}$}
  \ENDIF
  \STATE {$i = i+1$}
  \ENDWHILE
  \STATE $p_k = \dfrac{g(s_{\textbf{h}_k})}{\displaystyle\sum_l{g(s_{\textbf{h}_l})}}$ for ${\textbf{h}_k} \in \mathcal{H}$
  \WHILE{Motion planning fails}
  \STATE {Sample $\textit{h}$ according to \{$p_k$\}}
  \ENDWHILE
  
  \end{algorithmic}
  \label{alg:grasp_selection}
  \end{algorithm}
\end{footnotesize}
\end{wrapfigure}

Because the network generates one grasp for each point, there are numerous similar grasps in each grasp's neighborhood and we use non-maximum suppression (NMS) to select grasps $\textbf{h}$ with local maximum $s_{\textbf{h}_i}$ to generate executable grasp set $\mathcal H$. Then weighted random sampling is applied to sample one grasp to execute according to its grasp quality score.


\section{Experiments}
\subsection{Implementation Details}
 The input point cloud is first preprocessed, including workspace filtering, outliers removal, and voxel grid down-sampling. For training and validation, we sample $\frac18N$ points from the point set with viable grasps, $\frac78N$ from the remaining point set, and integrate them as the input of the network. For evaluation, we sample $N$ points at random from the preprocessed point cloud. $N$ is set to 25600 in our experiments. We implement our network in PyTorch, and train it using Adam~\citep{kingma2014adam} as the optimizer for 100 epochs with the initial learning rate $0.001$, which is decreased by $2$ every $20$ epochs.

\subsection{Superiority of $\mathbb{SE}(3)$ grasp}
\begin{wrapfigure}{R}{0.40\textwidth}
    \centering
    \setlength{\belowcaptionskip}{-0.5cm}
    \includegraphics[width=1\linewidth]{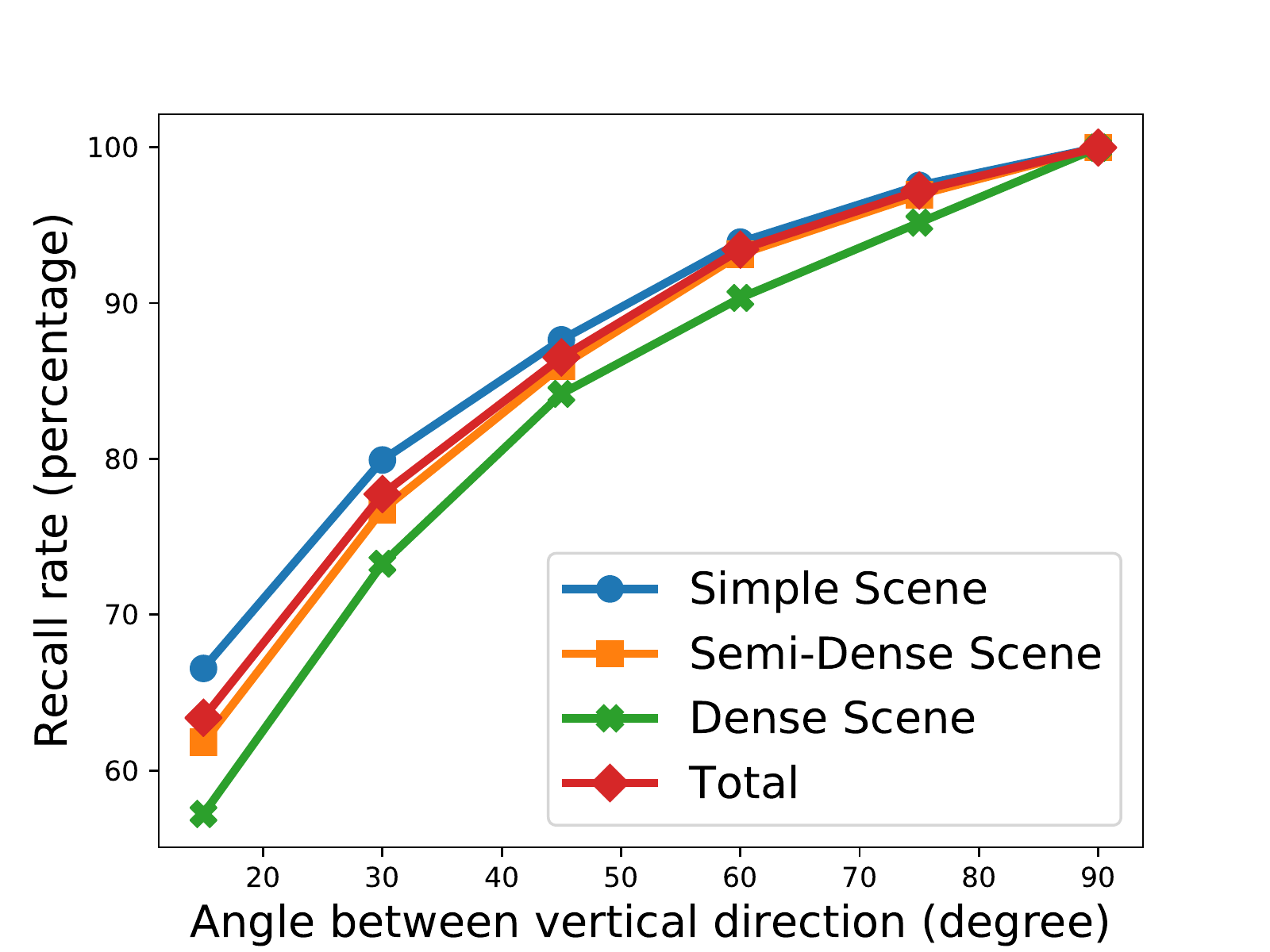}
    \caption{Results of simulated experiment on the recall rate of different grasp. The X-axis is the angle between the approach vector and vertical direction. The angle of absolute 3/4 DoF grasp is $0^{\circ}$}
    \label{fig:6dadvantage}
\end{wrapfigure}

We first evaluated the grasp quality performance of our proposed network on simulated data.
To demonstrate the superiority of $\mathbb{SE}(3)$ grasp over 3/4 DoF grasp, here we give a quantitative analysis over 6k scene with around 2.6M generated grasps (Fig. \ref{fig:6dadvantage}). In our experiments, grasps are uniformly divided into 6 groups according to the angle between the approach vector and vertical direction in the range of ($0^{\circ}, 90^{\circ}$). We use the recall rate as metric which are defined as the percentage of objects that can be grasped using grasps between vertical and certain angle. We evaluate the recall rate at scenes of three different densities: simple (1-5 objects presented in the scene), semi-dense (6-10 objects) and dense (11-15) objects. The overall recall rate is the weighted average of the three scenes. We find that only $63.38\%$ objects can be grasped by nearly vertical grasps ($0^{\circ}, 15^{\circ}$). With the increase of scene complexity, the advantage of $\mathbb{SE}(3)$ grasp becomes more remarkable.

{}

\subsection{Simulation Experiments}
\begin{table}[b]
\begin{footnotesize}
\setlength{\belowcaptionskip}{-0.5cm}
\vspace{-0.4cm}
\renewcommand\arraystretch{1.2}
    \centering
    \begin{tabular}{r|cc|cc}
        \toprule
        &\multicolumn{2}{c|}{\textbf{w/o Noise}} &\multicolumn{2}{c}{\textbf{ w/ Noise}} \\
        \hline
        & Antipodal Score    & Collision-free &   Antipodal Score    & Collision-free \\
        \hline
        GPD (3 channels)   &  0.5947 &  \textbf{47.07\%} &  0.5802  & 40.00\%    \\
        GPD (12 channels)   &  0.5883  &  45.27\%  &  0.5946  & 40.44\%    \\
        PointNetGPD    &  0.5718   &  42.41\%  & 0.6376   &  41.17\%   \\
        \hline
        Ours & \textbf{0.7364}    & 47.02\%  &\textbf{0.7354}  & \textbf{53.32\%} \\
        \bottomrule   
    \end{tabular}
    \caption{Comparison of grasp quality on simulation data.}
    \label{tab:simul}
\end{footnotesize}
\end{table}

GPD~\citep{ten2017grasp} and PointNetGPD~\citep{liang2019pointnetgpd} adopt Darboux frame analysis to sample grasp poses and train a classifier to evaluate their quality, which achieved state-of-the-art performance in 6D grasp detection. We choose GPD(3 channels), GPD(12 channels), and PointGPD as our baseline methods.
For training baseline methods, we adopt their grasp sampling strategy to generate grasp candidates for each scene until we get 300 collision-free grasps. We generate grasps over 6.5k scenes and get more than 2M grasps, which is larger than the 300K grasps in the original paper. Note that the scene used to generate training data for baseline method is exactly the same as our method.

For evaluation of baseline methods, we first sample 1000 points at random from the point cloud and calculate the Darboux frame for grasp candidates, which are then classified and ranked. The top 10 grasps are evaluated for both baseline methods and our methods.

To evaluate our method in finding collision-free grasps, we compare two metrics that affect the final grasping success rate: \textbf{(1)} antipodal score, which describes the force closure property of grasps, \textbf{(2)} probability of collision with other objects not observable to the depth sensor. The evaluation is performed in simulator with 2 settings: \textbf{(1)} No noise, where the point cloud from the depth sensor simulator aligns with the complete point cloud perfectly; \textbf{(2)} With noise, where the noise of the depth simulator is proportional to the depth. Please note that for noise setting and real-world experiment, both baselines and our method is trained on noisy data. Table.~\ref{tab:simul} shows the comparison results. Since the 6-DoF grasp pose is regressed by our \net~instead of being computed from local normals and curvatures, it is less sensitive to partial and noisy depth observations; also, our \net~is able to generate more collision-free grasps by inferring from local and global geometry information jointly.




\subsection{Robotic Experiments}

We validate the effectiveness and reliability of our methods in real robotic experiments. We carried out all the experiments on Kinova MOVO, a mobile manipulator with a Jaco2 arm attached with a 2-finger gripper (Fig.~\ref{fig:pipeline} (a)). In order to be close to real domestic robot application scenarios, we use one KinectV2 depth sensor that is mounted on the head of the manipulator, which makes the observation heavily occluded and raises the difficulty of experiments. 30 objects of various shapes and weight (see Supplymentary Materials) are used, which are absent in the training dataset. 

The experiment procedure is as follows: \textbf{(1)} Choose 10 out of the 30 objects at random and put them on the table to form a cluttered scene; \textbf{(2)} The robot attempts multiple grasps, until all objects are grasped or 15 grasps have been attempted; \textbf{(3)} Step (1) and (2) are repeated for 4 times for each method. More details are presented in the supplementary material. Note that all the objects selected in real robot experiments are out of the training data.

As illustrated in Table~\ref{tab:robot}, our method outperforms baseline methods in terms of \textit{success rate}, \textit{completion rate}, and time efficiency, which suggests that the single-shot regressed 6-DoF grasps have better force closure quality than sampled grasps from baselines, as demonstrated in Fig.~\ref{fig:sample_grasp}. Not needed by us, the baseline methods also need to detect collision and extract local geometry for every sampled grasp, which takes around 20 seconds, so they are much more time-consuming than our method. 

Our experiment setting is much more challenging than the baseline papers. In the original paper, GPD uses two depth sensors at both sides of the arena to capture the nearly complete point cloud in the original paper, but in our experiments, only one depth sensor is used. In both baselines, grasps are sampled in the neighbourhood of Darboux frame. It performs well on convex objects (box and ball) but poorly on non-convex or thin-structure objects, such as mug and bowl as in our experiments, because their heuristic sampling method requires accurate normals and curvatures but estimating those surface normals from noisy point cloud is challenging. On the contrary, Point-Net++ has been demonstrated to be robust against adversarial changes to the input data \cite{liu2019extending}, which can better capture the geometric structure under noise.

\begin{table}[]
\setlength{\belowcaptionskip}{-0.5cm}
    \centering
    \setlength{\belowcaptionskip}{-0.5cm}
    \setlength{\tabcolsep}{2.5mm}{
    \begin{tabular}{r|cc|ccc}
        \toprule
         &\multicolumn{2}{c|}{\textbf{Grasp quality}} &\multicolumn{3}{c}{\textbf{Time-efficiency}} \\
        \hline
            & Success rate  & Completion rate & Processing & Inference &\textbf{ Total}  \\
        \hline
        GPD (3 channels)   &  40.0\%  &  60.0\%  &  24106 ms   &  \textbf{1.50 ms}   & 24108 ms    \\
        GPD (12 channels)   &  33.3\%  & 50.0\%  & 27195ms  & 1.70ms &  27197ms \\
        PointNetGPD  &  40.0\%  &  60.0\%  &  17694ms  &  2.86ms    & 17697ms  \\
        \hline
        Ours& \textbf{77.1\%} & \textbf{92.5\%} & \textbf{5804ms} & 12.60 ms & \textbf{5817 ms}\\
        \bottomrule   
    \end{tabular}}
    \caption{Results of robotic experiments on dense clutters. \textit{Success rate} and \textit{completion rate} are used as the evaluation metrics, which represent the accuracy and completeness respectively. }
    \label{tab:robot}
\end{table}

\begin{figure}[b]
\setlength{\belowcaptionskip}{-0.5cm}
\vspace{-0.4cm}
\centering
    \includegraphics[width=1\linewidth]{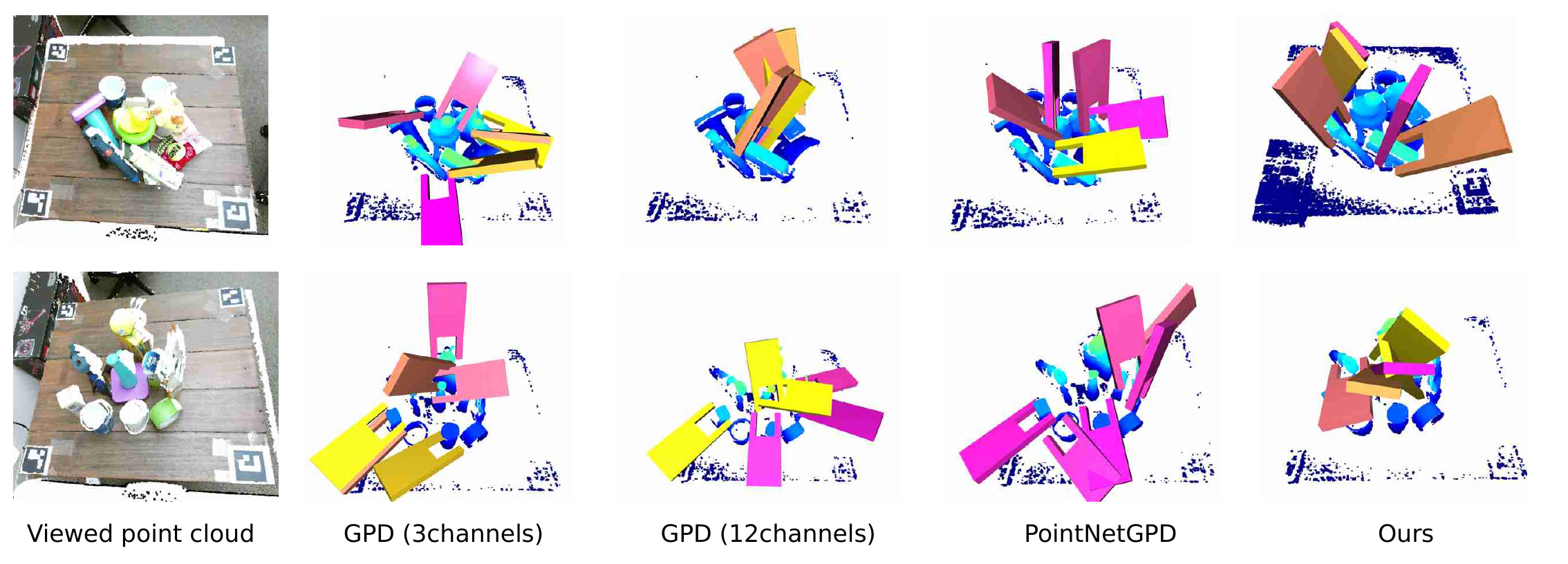}
    \caption{Comparison between sampled grasps chosen by baseline methods with high-score and regressed grasps by our method. }
    \label{fig:sample_grasp}
\end{figure}

\section{Conclusion}
We studied the problem of 6-DoF grasping by a parallel gripper in a cluttered scene captured using a commodity depth sensor from a single viewpoint. Our learning based approach trained in a synthetic scene can work well in real-world scenarios, with improved speed and success rate compared with state-of-the-arts. The success shows that our design choices, including a single-shot grasp proposal and a novel gripper contact model, are effective. 


\clearpage
\acknowledgments{We would like to acknowledge the National Science Founding for the grant RI-1764078 and Qualcomm for the generous support. We especially thank Jiayuan Gu for the discussion on network architecture design and Fanbo Xiang for the idea on using single object cache to accelerate training data generation.}


\bibliography{main}
\newpage
\appendix
\section{Supplementary Material}

\subsection{Network Details}
We use 3 point set abstract layers, each of which is a 3-layer MLP, containing $(128, 128, 256)$, $(256, 256, 512)$, $(512, 512, 1024)$ units, respectively. ReLU is used as the activation function. Farthest Point Sampling(FPS) is adopted for better and more uniform coverage, where a subset of points are chosen from the input point set such that each point in the subset is the most distant point from points in the set. Compared with random sampling, FPS has better coverage of the entire point set. It is performed iteratively to get the centroids for grouping from the former stage.

\subsection{Robotics Experiments Dataset}
\begin{figure}[h]
\centering
    \includegraphics[width=0.7\linewidth]{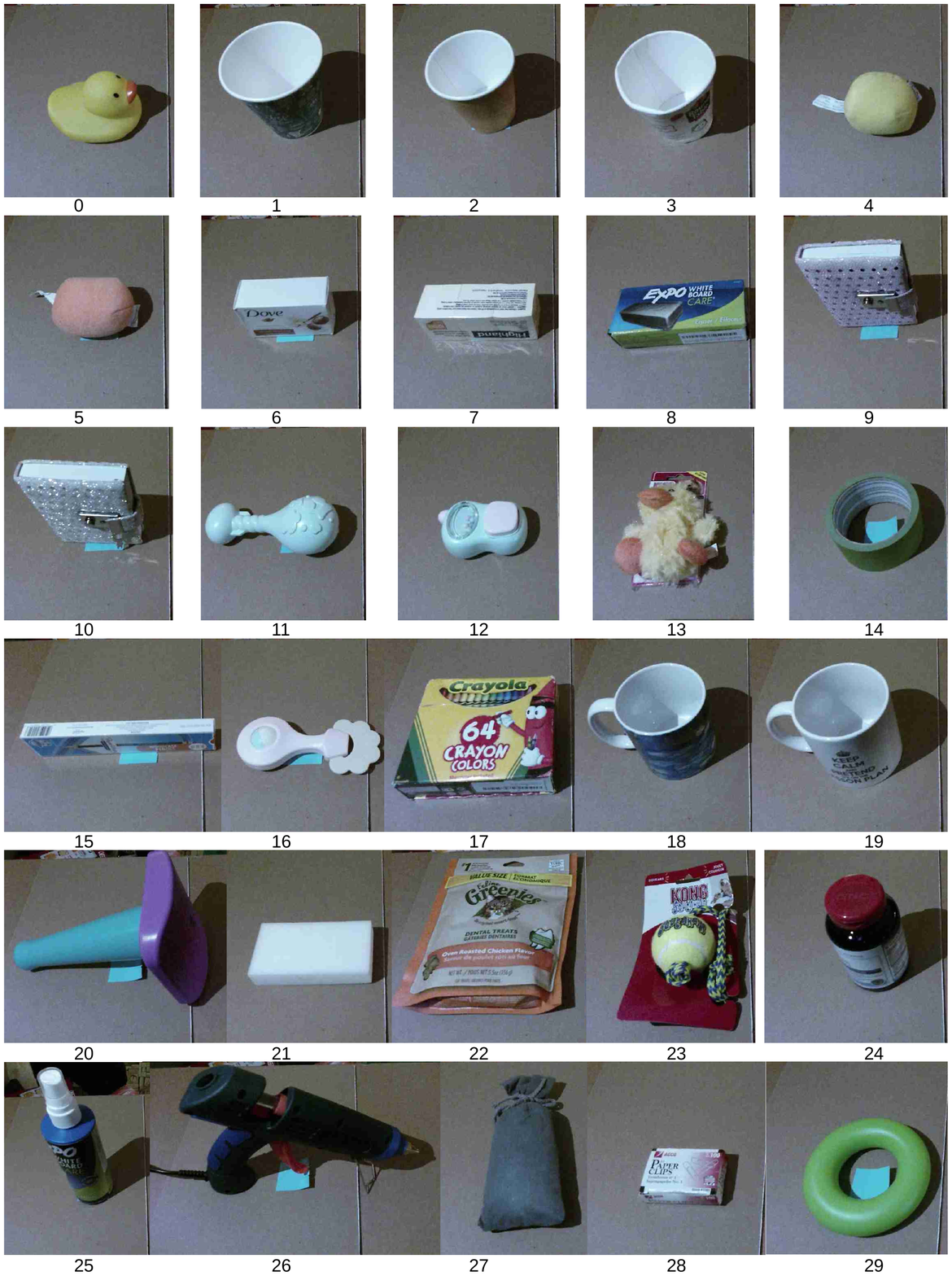}
    \caption{The 30 objects used in our experiments. }
    \label{fig:object}
\end{figure}

Figure.~\ref{fig:object} shows the 30 objects used in our experiments. This dataset is collected from daily objects and different from the YCB\citep{calli2015ycb} dataset we used to generate training data. 

\subsection{Robotics Experiments Grasp Proposal}

\begin{figure}[h]
\centering
    \includegraphics[width=1.0\linewidth]{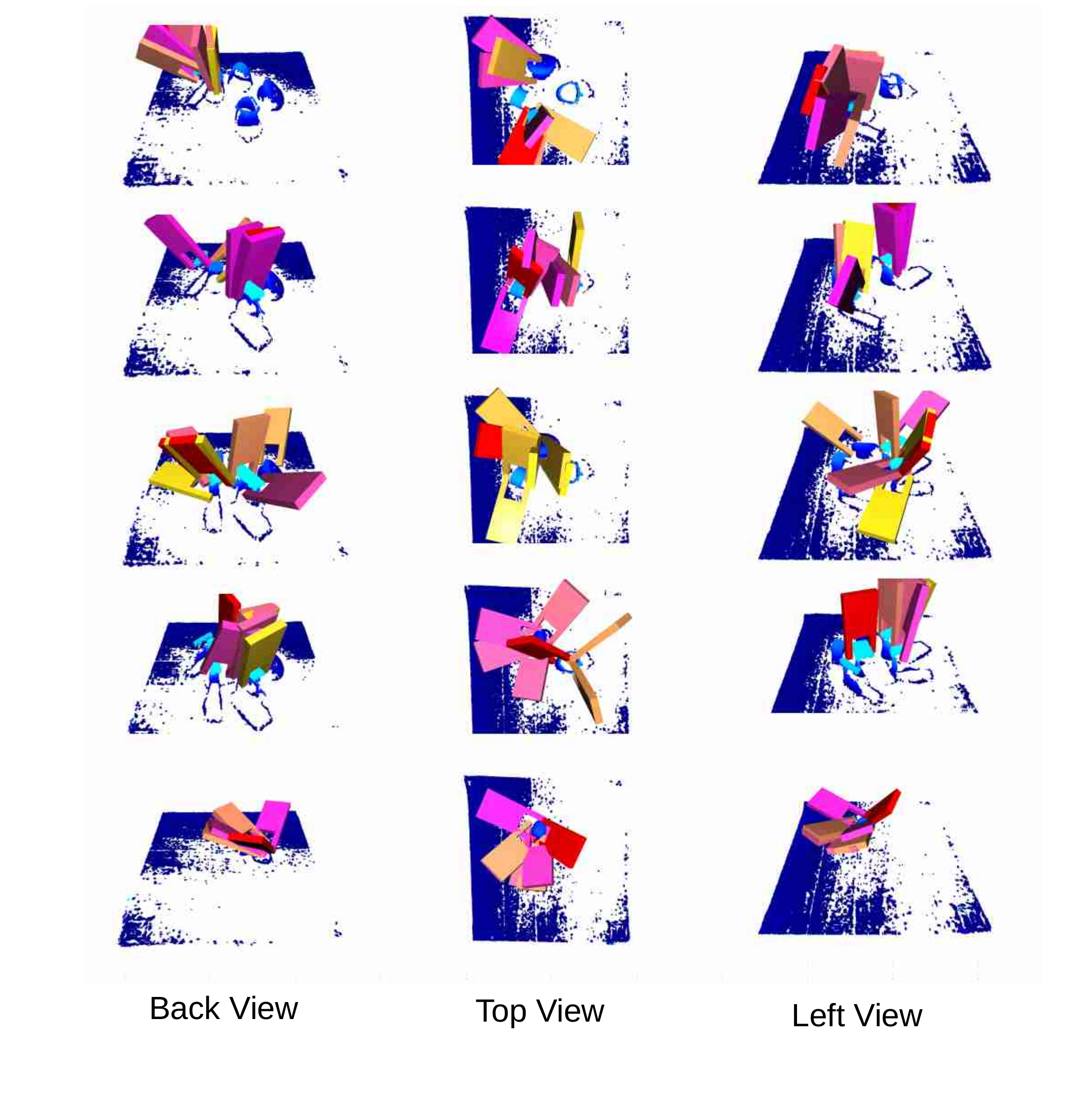}
    \caption{Viewed point cloud from the depth sensor and high quality grasp set in robotic experiments. }
    \label{fig:grasp}
\end{figure}

\begin{figure}[h]
\centering
    \includegraphics[width=1.0\linewidth]{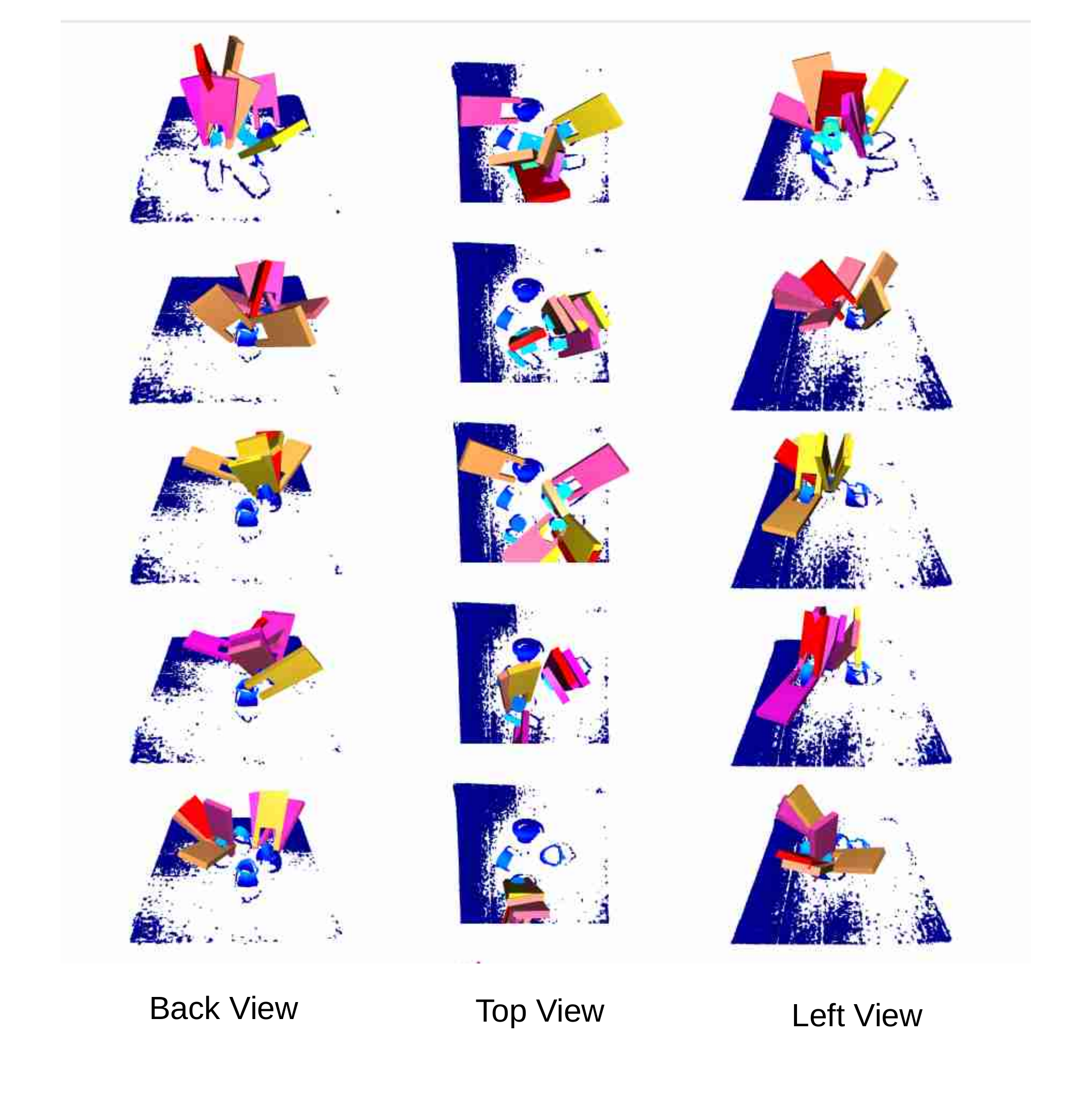}
    \caption{More viewed point cloud from the depth sensor and high quality grasp set in robotic experiments. }
    \label{fig:moregrasp}
\end{figure}

Figure.~\ref{fig:grasp}, ~\ref{fig:moregrasp} show the viewed point cloud and proposed high quality grasp set in robotic experiments.

\end{document}